\documentclass{article}
\usepackage[utf8]{inputenc}
\usepackage{url}
\usepackage{hyperref}
\usepackage[numbers, compress, sort]{natbib}
\usepackage{amsmath, amsfonts, amsthm}
\usepackage{cleveref}
\usepackage{mdframed}
\usepackage[shortlabels]{enumitem}
\usepackage{hyperref}
\usepackage{arxiv}
\usepackage{notation}
\usepackage{thmtools,thm-restate}
\usepackage{pgfplots}
\pgfplotsset{compat=1.15}
\usepackage{mathrsfs}
\usetikzlibrary{arrows}

\usetikzlibrary{snakes}

\crefname{figure}{Fig.}{Figs.}
\crefname{definition}{Defn.}{Defns.}
\crefname{corollary}{Corollary}{Corollaries}
\crefname{lemma}{Lemma}{Lemmas}
\crefname{proposition}{Prop.}{Props.}
\crefname{theorem}{Thm.}{Thms.}
\crefname{remark}{Remark}{Remarks}
\crefname{principle}{Principle}{Principles}
\crefname{lemma}{Lemma}{Lemmas}
\crefname{table}{Tab.}{Tabs.}
\crefname{section}{\S}{\S\S}
\crefname{subsection}{\S}{\S\S}
\crefname{subsubsection}{\S}{\S\S}

\title{On Pitfalls of Identifiability in Unsupervised Learning. A Note on: ``Desiderata for Representation Learning: A Causal Perspective''}
\author{Shubhangi Ghosh, Luigi Gresele, Julius von Kügelgen, Michel Besserve, Bernhard Schölkopf}
\date{November 2021}

\begin{document}

\maketitle

\begin{abstract}
Model identifiability is a desirable property %
in the context of unsupervised representation learning. \looseness=-1 In absence thereof, different models may be observationally indistinguishable %
while yielding representations that are nontrivially related to one another, thus making the recovery of %
a ground truth generative model fundamentally impossible---as often shown through suitably constructed counterexamples.
In this note, we discuss one such construction, illustrating a potential failure case of an identifiability result presented in {\em ``Desiderata for Representation Learning: A Causal Perspective''} by Wang \& Jordan (2021)~\cite{wang2021desiderata}.\footnote{Our note concerns v1 of the manuscript, Submitted on 8 Sep 2021.}
The construction is based on the theory of nonlinear independent component analysis. We comment on implications of this and other counterexamples for identifiable representation learning.
\end{abstract}

\section{Introduction}

One of the goals of representation learning is to infer latent structure of the data generating process starting from raw, unstructured observations. 
A way to formalise this is to postulate a generative model where the observations are given by a nonlinear transformation (also termed {\em mixing} function) of some latent variables of interest, and characterise under which conditions it is a priori possible to %
recover those ground truth variables---also termed {\em identifiability}.\footnote{see, e.g.,~\cite[][\S~2.3]{khemakhem2020variational} and~\cite[][\S~2.1]{Greseleetal21} for formal definitions in the context of representation learning and nonlinear ICA.} 
This has mainly been studied in the context of independent component analysis (ICA)~\cite{comon1994independent, ICAbook}, under the additional assumption that the latent variables of interest are statistically independent.
For the unsupervised i.i.d.~setting, a negative result states that recovering  the ground truth latent variables is fundamentally impossible~\cite{darmois1951construction, hyvarinen1999nonlinear}: without additional {\em auxiliary} variables~\cite{hyvarinen2019nonlinear} or restrictions on the function class to which the mixing belongs~\cite{Greseleetal21}, different models can be indistinguishable based on observations, but yield latent representations which differ in nontrivial ways.

In the related context of disentanglement, a recent paper, {\em ``Desiderata for Representation Learning: A Causal Perspective”}~\cite{wang2021desiderata}, presents a result (Theorem 11) discussing identifiability of latent, not necessarily independent, factors  in a fully unsupervised data-generating process. 
The main claim of the theorem is that, under suitable assumptions which we restate in~\cref{sec:existing_thm}, the latent factors are identifiable up to permutation and coordinate-wise transformations. 

This appears to be at odds with the aforementioned negative results for identifiability of nonlinear ICA in an unsupervised setting~\cite{hyvarinen1999nonlinear}. 
To illustrate this point, we provide a counterexample inspired by the theory of nonlinear ICA~(\cref{sec:proposed_counterexample}). In particular, we leverage the indeterminacy introduced by a measure-preserving automorphism presented in%
~\cite[][\S~2.2]{hyvarinen1999nonlinear}, to provide an instance where two representations satisfying the assumptions of Theorem~11 in~\cite{wang2021desiderata} are not coordinate-wise transformations and permutations of each other.

We stress that we claim no novelty regarding the construction presented in this note, which is based on theory of nonlinear ICA~\cite{hyvarinen1999nonlinear}. It should simply be thought of as exemplifying one among the many potential pitfalls that exist in identifiable representation learning.
We also acknowledge personal communication with the authors of~\cite{wang2021desiderata}, who kindly discussed this matter with us and acknowledged to upload a revised version of their paper taking this problem into account.

\section{Identifiability of representations with independent support - Theorem 11 in~\cite{wang2021desiderata}}
\label{sec:existing_thm}

\begin{theorem}[Identifiability of representations with independent support~\cite{wang2021desiderata}] \label{thm11-statement}
Among all compactly supported representations (i.e., the support being a closed and bounded region) that generate the same \(\sigma\)-algebra, the representation with \textit{independent support} (\ref{isupp}) (if it exists) is identifiable up to permutation and coordinate-wise bijective transformations.\\
\\
\textbf{Technical statement of the theorem}\\
Given:
\begin{itemize}
    \item observations \(\mathbf{X} \in \mathbb{R}^m\)
    \item latent representations \(\mathbf{Z} = \mathbf{f}(\mathbf{X}), \mathbf{Z'} = \mathbf{f'}(\mathbf{X}) \in \mathbb{R}^d \)
\end{itemize}
assuming the following, 

\begin{itemize}
    \item  \(\mathbf{f}, \mathbf{f'}\) are continuous
    \item  The sigma algebras of the latent representations\footnote{Definition of \(\sigma\)-algebra generated by a random variable: \href{https://https://en.wikipedia.org/wiki/\%CE\%A3\-algebra\#\%CF\%83\-algebra\_generated\_by\_random\_variable\_or\_vector}{\textcolor{blue}{link}}} are the same, i.e., \(\sigma(\mathbf{Z}) = \sigma(\mathbf{Z'})\) 
    
    \item \(\mathbf{Z}, \mathbf{Z'}\) both have compact support in \(\mathbb{R}^d\)
    \item \(\mathbf{Z}, \mathbf{Z'}\) both satisfy the \textit{independent support condition} (\ref{isupp})
\end{itemize}
we have,
\begin{equation}
    Z_1, Z_2, \ldots Z_d = \text{perm}(q_1(Z'_1), q_2(Z'_2), \ldots, q_d(Z'_d)), \label{thm11}
\end{equation}
where \(q_j\) are continuous bijective functions with a compact domain in \(\mathbb{R}\).

\end{theorem}

We also restate the definition of the {\em independent support condition}, which is important in the context of~\cref{thm11-statement}.

\begin{definition}[Independent Support Condition~\cite{wang2021desiderata}]
\label{isupp}
    For a representation \(\mathbf{Z} = (Z_1, Z_2, \ldots, Z_d)\), the independent support condition is satisfied if 
\begin{equation}
    supp(Z_1, Z_2, \ldots, Z_d) = supp(Z_1) \times supp(Z_2) \times \ldots \times supp(Z_d) 
\end{equation}
\end{definition}

\section{Proposed counterexample}
\label{sec:proposed_counterexample}
We take inspiration from a setting of nonlinear ICA in two dimensions (see~\cite[][\S~2.2]{hyvarinen1999nonlinear}) to construct a counterexample to%
~\cref{thm11-statement}.

\begin{itemize}
    \item Consider latent factors \(\Zb \sim \text{Unif}\left([-1, 1]^2\right)\). 
    \item Consider the mapping of the latent factors to the observations, \(\Xb = A\Zb\) where \(A \in \RR^{2 \times 2}\) is an invertible matrix.\footnote{Note that a nonlinear invertible map $\fb$ could also be used here in place of the linear map $A$.} The probability density of \(\Xb\) is supported on a compact parallelepiped in \(\RR^2\). Note that \(\Zb = \fb(\Xb)\) where \(\fb \equiv A^{-1}\).
    \item We construct an alternate latent representation \(\Zb'= \hb(\Zb)\) defined as a transformation of \(\Zb\) as follows:
    \begin{equation}
    \mathbf{h}(\mathbf{Z}) = \begin{cases}
                            \mathbf{Z}\,, & \|\mathbf{Z}\| > c \\
                            \mathbf{Z}\exp(ia(\|\mathbf{Z}\| - c))\,, & \|\mathbf{Z}\| \le c
                        \end{cases} \label{eq:mpa-simple}
    \end{equation}

    where \(c\) is a real number in \((0, 1)\), \(i\) is the imaginary unit, and \(a\) is a nonzero real number.

    Note that the probability density of \(\Zb'\) is also supported on \([-1, 1]^2\). \footnote{It can also be shown that \(\Zb' \sim \text{Unif}\left([-1, 1]^2\right)\).} \(\Zb' = \fb'(\Xb)\), where \(\fb' \equiv \hb \circ \fb\)
    
\end{itemize}

Thereby, we have constructed continuous bijective mappings from random observations \(\Xb\) to two distinct latent representations \(\Zb\) and \(\Zb'\) supported on \([-1, 1]^2\), a compact subset of \(\RR^2\).

\begin{figure}
    \centering
    \includegraphics[scale=0.7]{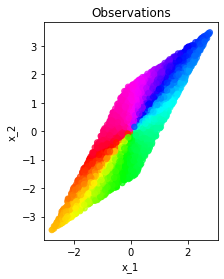}\hfill\\
    \includegraphics[scale=0.6]{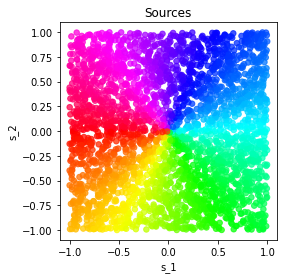}
    \includegraphics[scale=0.6]{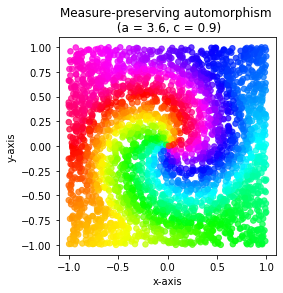}
    \caption{Observations and latent representations generated as described in~\cref{sec:proposed_counterexample}.
    \textbf{Top:} Observed variables; \textbf{Bottom, left:} Sources $\Zb$; \textbf{Bottom, right:} Sources $\Zb'$ generated as in~\eqref{eq:mpa-simple}, with parameters $a=3.6$ and $c=0.9$.}
    \label{fig:measure_preserving_automorphism}
\end{figure}

The indeterminacy introduced by \(\mathbf{h}\) is not trivial. In particular, it cannot be generally expressed in terms of coordinate-wise transformations and permutations as postulated by Theorem 11 (\ref{thm11-statement}). For an example choice of parameters, we include a visualisation of the nonlinear distortion between $\Zb$ and $\Zb'$ in~\cref{fig:measure_preserving_automorphism}. 

In order to show that the construction is a valid counterexample,
we need to verify that it satisfies the assumptions of~\cref{thm11-statement}.

\subsection{Verifying that the proposed counterexample satisfies the assumptions of~\cref{thm11-statement}} \label{verify-simple}

\begin{itemize}
    \item \textbf{Continuity of \(\fb\) and \(\fb'\)}:
    
    \(\fb \equiv A^{-1}\) is continuous since it is a linear map. 
    
    \(\hb(\cdot)\) is continuous by definition (see also \cite{hyvarinen1999nonlinear}).

    \(\fb' \equiv \hb \circ \fb\) is the composition of two functions which are continuous at all points, and hence is also continuous (see, e.g.,~\cite[][Thm.~4.7]{rudin1964principles}).

    \item \textbf{\(\sigma(\mathbf{Z}) = \sigma(\mathbf{Z'})\)}:

    We show that \(\hb(\cdot)\) is (continuous) bijective with continuous inverse. First, we note that this is a radius- preserving function, such that we  only need to check bijectivity with domain and co-domain restricted to each fixed radius. In the case \(\|\mathbf{Z}\| > c\), \(\hb(\cdot)\) is the identity map and hence is bijective. In the case \(\|\mathbf{Z}\| \le c\), \(\hb(\cdot)\) is a radius-dependent rotation and hence is also bijective, with inverse function obtained by replacing $a$ with $-a$ in~\cref{eq:mpa-simple}. This last observation shows that the inverse is continuous. 
    
    As \(\Zb\) and \(\Zb'\) can be written as a continuous function of each other in both ways, \(\Zb = \hb(\Zb')\) and \(\Zb' = \hb^{-1}(\Zb)\), this implies \(\sigma(\mathbf{Z}) = \sigma(\mathbf{Z'})\) (see~\cite[][\S~3.3]{wang2021desiderata}).
    
    \item \textbf{Compactness of support}:
    \(\mathbf{Z}, \mathbf{Z'}\) are both supported in \([-1,1]^2\), which is a compact region in~\(\mathbb{R}^2\).
    
    \item \textbf{Checking Independent Support Condition}:
        Both \(\mathbf{Z}\) and \(\mathbf{Z'}\) have the (hyper-)rectangular support \([-1,1]^2\), hence satisfy the Independent Support Condition (\ref{isupp}).

\end{itemize}

Hence the proposed counterexample satisfies the premises of~\cref{thm11}. However, \(\mathbf{Z}\) and \(\mathbf{Z'}\) are related by the measure-preserving automorphism \(\mathbf{h}\)%
, which in general cannot be expressed as a permutation and coordinate-wise bijection transformation. %

\section{Conclusion}
The construction in~\cref{sec:proposed_counterexample} allows us to build a counterexample which satisfies the assumptions in~\cref{thm11-statement}, in particular the {\em independent support condition}.

Other counterexamples can be given to illustrate the identifiability issues present in nonlinear ICA and disentanglement~\cite{hyvarinen1999nonlinear, locatello2019challenging}. While most of these concern settings where the latent variables are assumed to be independent, they also bear broader implications: as argued in~\cite[][Appendix~D]{khemakhem2020variational}, in the unsupervised, i.i.d. setting, {\em ``models with {\it any} form of unconditional prior [...] are unidentifiable''}.

At the same time, while `ruling out' families of counterexamples is a necessary condition for identifiability, it is not a sufficient one: for example, our recent work~\cite{Greseleetal21} rules out some classes of counterexamples typically used in nonlinear ICA and disentanglement by imposing constraints on the function class to which the mixing belongs.
This can, however, not yet be considered as a full identifiability result but only a first step in that direction.

A different line of work tries to recover identifiability by postulating the availability of additional information, such as auxiliary variables~\cite{hyvarinen2019nonlinear, khemakhem2020variational, hyvarinen2016unsupervised, hyvarinen2017nonlinear, roeder2021linear}, temporal or spatial structure in the data data~\cite{halva2021disentangling} or multiple `views'~\cite{gresele2020incomplete, locatello2020weakly, zimmermann2021contrastive}.
The above works are also limited to the assumption of (conditionally) independent latent variables; suitable notions of identifiability may however also be provided for settings where the latent variables are linked via nontrivial causal relationships, see for example~\cite{von2021self}. %

The authors of~\cite{wang2021desiderata} have meanwhile uploaded a new version of their work~\cite{wang2022desiderata} which introduces an additional assumption to address the issue we raised. The additional assumption %
(see assumption 2 in Theorem 11 in the updated version~\cite{wang2022desiderata}, uploaded on the 10th of February 2022) constrains the relationship between latent factors in two alternative representations $\Zb$ and $\Zb'$.  %
As we argued above, assumptions for identifiability are instead typically phrased as conditions on the data generating process (e.g. as constraints on the mixing function). %
It is currently unclear how the new assumption should be interpreted in terms of the data generating process and to what extent it entails testable implications; we believe that more work may be needed to elaborate on this.

\section*{Acknowledgements}
We thank Yixin Wang for discussions about a first version of the present note. We also thank Aapo Hyv\"arinen, Damien Teney and Francesco Locatello for helpful discussions.

{\small
\bibliographystyle{plainnat}
\bibliography{ref}

\begin{thebibliography}{19}
\providecommand{\natexlab}[1]{#1}
\providecommand{\url}[1]{\texttt{#1}}
\expandafter\ifx\csname urlstyle\endcsname\relax
  \providecommand{\doi}[1]{doi: #1}\else
  \providecommand{\doi}{doi: \begingroup \urlstyle{rm}\Url}\fi

\bibitem[Comon(1994)]{comon1994independent}
Pierre Comon.
\newblock Independent component analysis, a new concept?
\newblock \emph{Signal processing}, 36\penalty0 (3):\penalty0 287--314, 1994.

\bibitem[Darmois(1951)]{darmois1951construction}
George Darmois.
\newblock Analyse des liaisons de probabilit{\'e}.
\newblock In \emph{Proc. Int. Stat. Conferences 1947}, page 231, 1951.

\bibitem[Gresele et~al.(2021)Gresele, von K{\"u}gelgen, Stimper, Sch{\"o}lkopf,
  and Besserve]{Greseleetal21}
L.~Gresele, J.~von K{\"u}gelgen, V.~Stimper, B.~Sch{\"o}lkopf, and M.~Besserve.
\newblock Independent mechanism analysis, a new concept?
\newblock In \emph{Advances in Neural Information Processing Systems 34
  (NeurIPS 2021)}, December 2021.

\bibitem[Gresele et~al.(2019)Gresele, Rubenstein, Mehrjou, Locatello, and
  Sch{\"o}lkopf]{gresele2020incomplete}
Luigi Gresele, Paul~K Rubenstein, Arash Mehrjou, Francesco Locatello, and
  Bernhard Sch{\"o}lkopf.
\newblock The {I}ncomplete {R}osetta {S}tone problem: Identifiability results
  for multi-view nonlinear {ICA}.
\newblock In \emph{Uncertainty in Artificial Intelligence}, pages 217--227.
  PMLR, 2019.

\bibitem[H{\"a}lv{\"a} et~al.(2021)H{\"a}lv{\"a}, Le~Corff, Leh{\'e}ricy, So,
  Zhu, Gassiat, and Hyvarinen]{halva2021disentangling}
Hermanni H{\"a}lv{\"a}, Sylvain Le~Corff, Luc Leh{\'e}ricy, Jonathan So,
  Yongjie Zhu, Elisabeth Gassiat, and Aapo Hyvarinen.
\newblock Disentangling identifiable features from noisy data with structured
  nonlinear {ICA}.
\newblock \emph{Advances in Neural Information Processing Systems}, 34, 2021.

\bibitem[Hyvarinen and Morioka(2016)]{hyvarinen2016unsupervised}
Aapo Hyvarinen and Hiroshi Morioka.
\newblock Unsupervised feature extraction by time-contrastive learning and
  nonlinear ica.
\newblock \emph{Advances in Neural Information Processing Systems},
  29:\penalty0 3765--3773, 2016.

\bibitem[Hyvarinen and Morioka(2017)]{hyvarinen2017nonlinear}
Aapo Hyvarinen and Hiroshi Morioka.
\newblock Nonlinear ica of temporally dependent stationary sources.
\newblock In \emph{Artificial Intelligence and Statistics}, pages 460--469.
  PMLR, 2017.

\bibitem[Hyv{\"a}rinen and Pajunen(1999)]{hyvarinen1999nonlinear}
Aapo Hyv{\"a}rinen and Petteri Pajunen.
\newblock Nonlinear independent component analysis: Existence and uniqueness
  results.
\newblock \emph{Neural networks}, 12\penalty0 (3):\penalty0 429--439, 1999.

\bibitem[Hyv{\"a}rinen et~al.(2001)Hyv{\"a}rinen, Karhunen, and Oja]{ICAbook}
Aapo Hyv{\"a}rinen, Juha Karhunen, and Erkki Oja.
\newblock \emph{Independent {C}omponent {A}nalysis}.
\newblock John Wiley \& Sons, Ltd, 2001.

\bibitem[Hyv{\"a}rinen et~al.(2019)Hyv{\"a}rinen, Sasaki, and
  Turner]{hyvarinen2019nonlinear}
Aapo Hyv{\"a}rinen, Hiroaki Sasaki, and Richard Turner.
\newblock Nonlinear {ICA} using auxiliary variables and generalized contrastive
  learning.
\newblock In \emph{The 22nd International Conference on Artificial Intelligence
  and Statistics}, pages 859--868. PMLR, 2019.

\bibitem[Khemakhem et~al.(2020)Khemakhem, Kingma, Monti, and
  Hyvarinen]{khemakhem2020variational}
Ilyes Khemakhem, Diederik Kingma, Ricardo Monti, and Aapo Hyvarinen.
\newblock Variational autoencoders and nonlinear ica: A unifying framework.
\newblock In \emph{International Conference on Artificial Intelligence and
  Statistics}, pages 2207--2217. PMLR, 2020.

\bibitem[Locatello et~al.(2019)Locatello, Bauer, Lucic, Raetsch, Gelly,
  Sch{\"o}lkopf, and Bachem]{locatello2019challenging}
Francesco Locatello, Stefan Bauer, Mario Lucic, Gunnar Raetsch, Sylvain Gelly,
  Bernhard Sch{\"o}lkopf, and Olivier Bachem.
\newblock Challenging common assumptions in the unsupervised learning of
  disentangled representations.
\newblock In \emph{international conference on machine learning}, pages
  4114--4124. PMLR, 2019.

\bibitem[Locatello et~al.(2020)Locatello, Poole, R{\"a}tsch, Sch{\"o}lkopf,
  Bachem, and Tschannen]{locatello2020weakly}
Francesco Locatello, Ben Poole, Gunnar R{\"a}tsch, Bernhard Sch{\"o}lkopf,
  Olivier Bachem, and Michael Tschannen.
\newblock Weakly-supervised disentanglement without compromises.
\newblock In \emph{International Conference on Machine Learning}, pages
  6348--6359. PMLR, 2020.

\bibitem[Roeder et~al.(2021)Roeder, Metz, and Kingma]{roeder2021linear}
Geoffrey Roeder, Luke Metz, and Durk Kingma.
\newblock On linear identifiability of learned representations.
\newblock In \emph{International Conference on Machine Learning}, pages
  9030--9039. PMLR, 2021.

\bibitem[Rudin et~al.(1964)]{rudin1964principles}
Walter Rudin et~al.
\newblock \emph{Principles of mathematical analysis}, volume~3.
\newblock McGraw-hill New York, 1964.

\bibitem[von K{\"u}gelgen et~al.(2021)von K{\"u}gelgen, Sharma, Gresele,
  Brendel, Sch{\"o}lkopf, Besserve, and Locatello]{von2021self}
Julius von K{\"u}gelgen, Yash Sharma, Luigi Gresele, Wieland Brendel, Bernhard
  Sch{\"o}lkopf, Michel Besserve, and Francesco Locatello.
\newblock Self-supervised learning with data augmentations provably isolates
  content from style.
\newblock In \emph{Advances in Neural Information Processing Systems}, 2021.

\bibitem[Wang and Jordan(2021)]{wang2021desiderata}
Yixin Wang and Michael~I Jordan.
\newblock Desiderata for representation learning: A causal perspective.
\newblock \emph{arXiv preprint arXiv:2109.03795v1}, 2021.
\newblock Our note concerns v1 of the manuscript, Submitted on 8 Sep 2021.

\bibitem[Wang and Jordan(2022)]{wang2022desiderata}
Yixin Wang and Michael~I. Jordan.
\newblock Desiderata for representation learning: A causal perspective.
\newblock \emph{arXiv preprint arXiv:2109.03795v2}, 2022.
\newblock Updated version was submitted on 10 Feb 2022.

\bibitem[Zimmermann et~al.(2021)Zimmermann, Sharma, Schneider, Bethge, and
  Brendel]{zimmermann2021contrastive}
Roland~S Zimmermann, Yash Sharma, Steffen Schneider, Matthias Bethge, and
  Wieland Brendel.
\newblock Contrastive learning inverts the data generating process.
\newblock In \emph{International Conference on Machine Learning}, pages
  12979--12990. PMLR, 2021.

\end{thebibliography}
}

\end{document}